\documentclass[letterpaper]{article} 
\usepackage{aaai23}  
\usepackage{times}  
\usepackage{helvet}  
\usepackage{courier}  
\usepackage[hyphens]{url}  
\usepackage{graphicx} 
\usepackage{natbib}  
\usepackage{caption} 
\usepackage{algorithm}
\usepackage{algorithmic}

\usepackage{newfloat}
\usepackage{listings}
\DeclareCaptionStyle{ruled}{labelfont=normalfont,labelsep=colon,strut=off} 
\floatstyle{ruled}
\newfloat{listing}{tb}{lst}{}
\floatname{listing}{Listing}
\usepackage{subfiles}
\usepackage{multirow}
\usepackage{verbatim}
\usepackage{graphicx}
\usepackage{amsmath}
\usepackage{amssymb}
\usepackage{wrapfig}
\usepackage{xcolor}
\usepackage{nicefrac}       
\usepackage{booktabs}

\title{Actional Atomic-Concept Learning for Demystifying\\ Vision-Language Navigation}
\author{
Bingqian Lin\textsuperscript{\rm 1\thanks{Part of this work was done during an internship in Huawei Noah's Ark Lab.}},
Yi Zhu\textsuperscript{\rm 2},
Xiaodan Liang\textsuperscript{\rm 1,3\thanks{Corresponding author.}
},
Liang Lin\textsuperscript{\rm 4},
Jianzhuang Liu\textsuperscript{\rm 2}
}
\affiliations{
\textsuperscript{\rm 1}Shenzhen Campus of Sun Yat-sen University, Shenzhen, 
\textsuperscript{\rm 2}Huawei Noah's Ark Lab\\
\textsuperscript{\rm 3}PengCheng Laboratory, \textsuperscript{\rm 4}Sun Yat-sen University\\
linbq6@mail2.sysu.edu.cn, 
zhuyi36@huawei.com, liangxd9@mail.sysu.edu.cn,\\
linliang@ieee.org, 
liu.jianzhuang@huawei.com
}

\usepackage{bibentry}

\begin{document}

\maketitle

\begin{abstract}
Vision-Language Navigation (VLN) is a challenging task which requires an agent to align complex visual observations to language instructions to reach the goal position. Most existing VLN agents directly learn to align the raw directional features and visual features trained using one-hot labels to linguistic instruction features. However, the big semantic gap among these multi-modal inputs makes the alignment difficult and therefore limits the navigation performance. In this paper, we propose Actional Atomic-Concept Learning (AACL), which maps visual observations to actional atomic concepts for facilitating the alignment. Specifically, an actional atomic concept is a natural language phrase containing an atomic action and an object, e.g., ``go up stairs''. These actional atomic concepts, which serve as the bridge between observations and instructions, can effectively mitigate the semantic gap and simplify the alignment. AACL contains three core components: 1) a concept mapping module to map the observations to the actional atomic concept representations through the VLN environment and the recently proposed Contrastive Language-Image Pretraining (CLIP) model, 2) a concept refining adapter to encourage more instruction-oriented object concept extraction by re-ranking the predicted object concepts by CLIP, and 3) an observation co-embedding module which utilizes concept representations to regularize the observation representations. Our AACL establishes new state-of-the-art results on both fine-grained (R2R) and high-level (REVERIE and R2R-Last) VLN benchmarks. Moreover, the visualization shows that AACL significantly improves the interpretability in action decision. 

\end{abstract}

\begin{figure*}[t]
\begin{centering}
\includegraphics[width=0.98\linewidth]{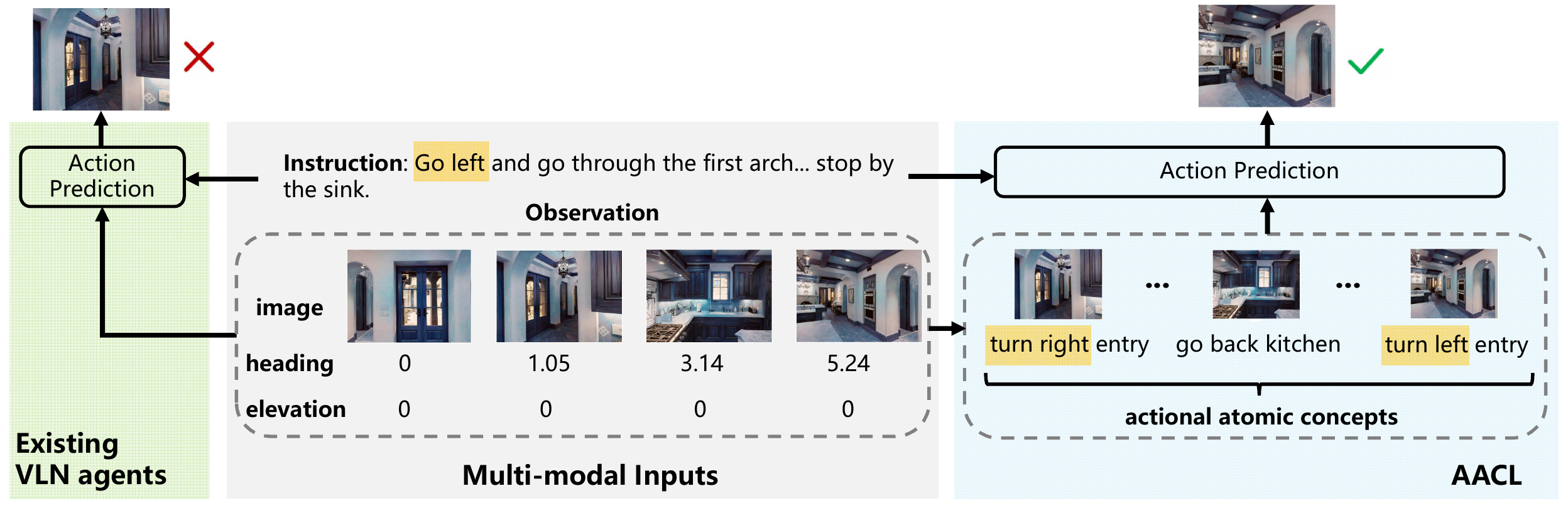}
\par\end{centering}
\caption{Comparison between existing VLN agents and the proposed AACL. Through mapping the visual observations to actional atomic concepts, AACL can simplify the multi-modal alignment and distinguish different observation candidates easily to make accurate action decision. 
}
\vspace{-0.6cm}
\label{fig:motivation}
\end{figure*}

\section{Introduction}
Vision-Language Navigation (VLN)~\cite{anderson2018vision,ku2020room,Chen2019TOUCHDOWNNL,Nguyen2019HelpAV} has attracted increasing interests in robotic applications since an instruction-following navigation agent is  practical and flexible in real-world scenarios. 
For accomplishing successful navigation, a VLN agent needs to align complicated visual observations to language instructions to reach the required target point. For example, when asking to ``turn left to the bathroom'', the agent should choose the right observation which not only contains the mentioned object ``bathroom'' but also indicates the direction ``turn left''.

Most of early VLN approaches adopt the LSTM-based encoder-decoder framework~\cite{fried2018speaker,tan2019learning,wang2019reinforced,ma2019self,zhu2020vision}, which encodes both the visual observations and language instructions and then generates the action sequence. With the development of large-scale cross-modal pretraining models in vision-language tasks~\cite{li2020unicoder,li2020oscar,chen2020uniter,lu2019vilbert}, emerging works attempt to introduce them into VLN tasks~\cite{hao2020towards,hong2021vln,Chen2021HistoryAM,Moudgil2021SOATAS}. 
However, both the non-pretraining-based or pretraining-based approaches represent the visual observations by raw directional features and visual features trained using one-hot labels, which are difficult to be aligned to the linguistic instruction features due to the large semantic gap among them. This direct alignment process also leads to poor interpretability of action decision and therefore makes the VLN agents unreliable to be deployed to real environments.


In this work, we aim to {\it mitigate the semantic gap and simplify the alignment} in VLN by proposing a new framework, called Actional Atomic-Concept Learning (AACL). Since the instructions usually consist of atomic action concepts, e.g., ``turn right'', and object concepts\footnote{In this work, we also treat the scene concept, e.g., ``bathroom'', as the object concept. }, e.g., ``kitchen'', in AACL, the visual observations are mapped to actional atomic concepts, which are natural language phrases each containing an action and an object. The actions are extracted from a predefined atomic action set. These actional atomic concepts, which can be viewed as the bridge between observations and instructions,  effectively facilitate the alignment as well as provide good interpretability for action decision. 

AACL consists of three main components. Firstly, a \textbf{concept mapping module} is constructed to map each single view observation to the actional atomic concept. For deriving the object concept, we resort to the recently proposed Contrastive Language-Image Pretraining (CLIP) model~\cite{radford2021learning} rather than image classification or object detection models pretrained on a fixed category set. Benefiting from the powerful open-world object recognition ability of CLIP, AACL can better generalize to diverse navigation scenarios. And we map the sequential direction information in VLN environments during navigation to the action concept.  Secondly, to encourage more instruction-oriented object concept extraction for facilitating the multi-modal alignment, a \textbf{concept refining adapter} is further introduced  to re-rank the predicted object concepts of CLIP according to the instruction. 
Lastly, an  \textbf{observation co-embedding module} embeds each observation and its paired actional atomic concept, and then uses concept representations to regularize the observation representations through an observation contrast strategy. Figure~\ref{fig:motivation} presents an action selection comparison between existing VLN agents and our AACL. Through mapping visual observations to actional atomic concepts formed by language, AACL can simplify the modality alignment and distinguish different action candidates easier to make correct actions.

We conduct experiments on several popular VLN benchmarks, including one with fine-grained instructions (R2R~\cite{anderson2018vision}) and two with high-level instructions (REVERIE~\cite{qi2020reverie} and R2R-Last~\cite{Chen2021HistoryAM}). Experimental results show that our AACL outperforms the state-of-the-art approaches on all benchmarks. Moreover, benefiting from these actional atomic concepts, AACL shows excellent interpretability in making action decision, which is a step closer towards developing reliable VLN agents in real-world applications.

\begin{figure*}[t]
\begin{centering}
\includegraphics[width=0.98\linewidth]{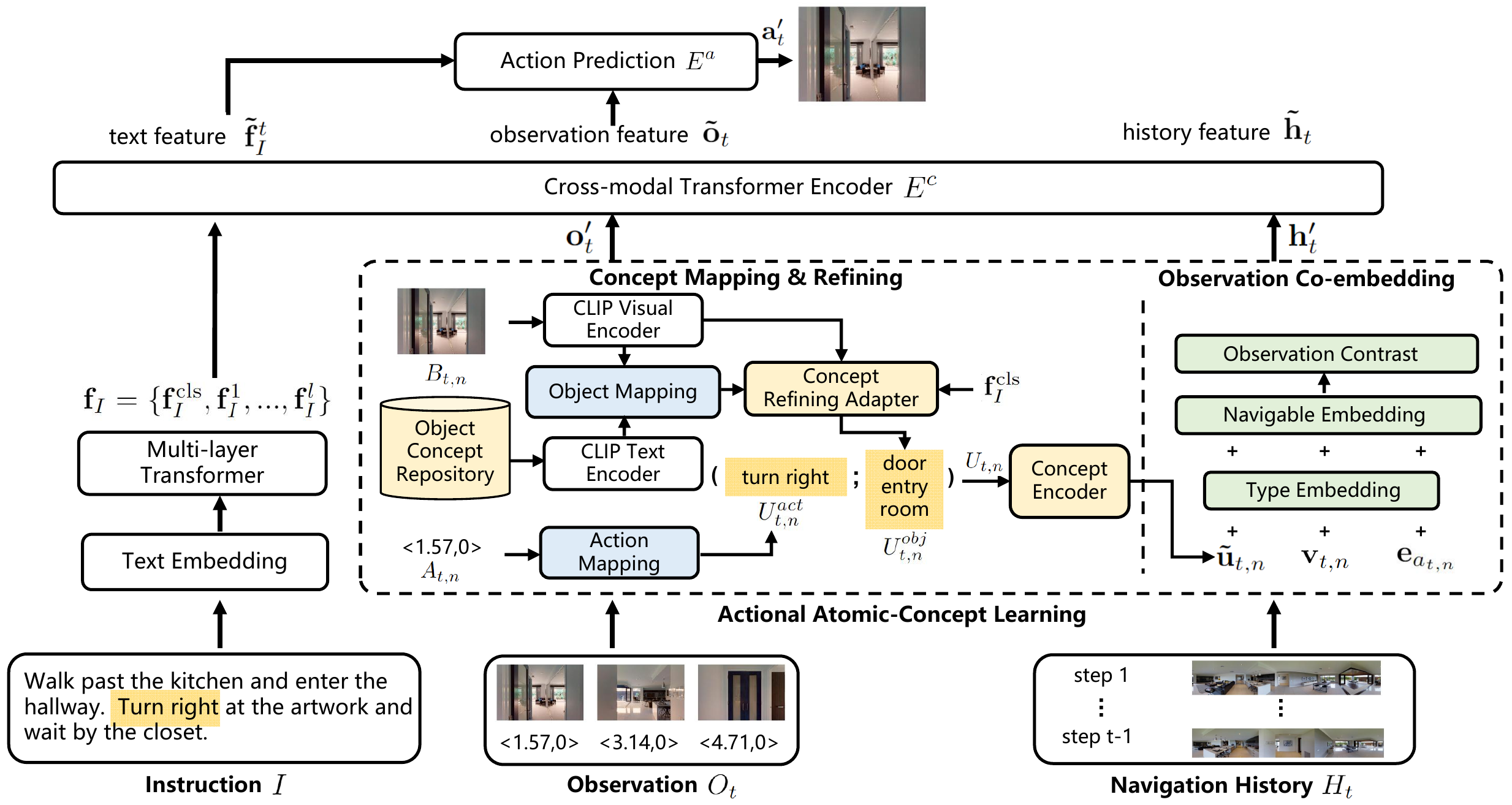}
\par\end{centering}
\vspace{-0.2cm}
\caption{Overview of our Actional Atomic-Concept Learning (AACL). At each timestep $t$, the agent receives the instruction $I$, the observation $O_{t}$, and the navigation history $H_{t}$. For each $O_{t,n}$ in $O_{t}$ containing the single-view image $B_{t,n}$ and the direction $A_{t,n}$,  object concept mapping and action concept mapping are conducted based on the concept refining adapter to obtain the actional atomic concept representations $\mathbf{\tilde{u}}_{t,n}$. Then $\mathbf{\tilde{u}}_{t,n}$ is used to regularize the visual representation $\mathbf{v}_{t,n}$ and the directional representation $\mathbf{e}_{A_{t,n}}$ through the observation co-embedding module for making action selection. For simplicity, we omit the learning process of navigation histories $H_{t}$, which is similar to that of observations $O_{t}$. 
}
\vspace{-0.6cm}
\label{fig:overview}
\end{figure*}

\section{Related Work}
\textbf{Vision-Language Navigation.}
Developing navigation agents which can follow natural language instructions has attracted increasingly research interests in recent years~\cite{anderson2018vision,ku2020room,Chen2019TOUCHDOWNNL,Nguyen2019HelpAV,qi2020reverie}. Most of early Vision-Language Navigation (VLN) approaches employ the LSTM-based encoder-decoder framework~\cite{fried2018speaker,tan2019learning,wang2019reinforced,ma2019self,zhu2020vision,Wang2020SoftER,Qi2020ObjectandActionAM,fu2020counterfactual}. Speaker-Follower~\cite{fried2018speaker} introduces a novel instruction augmentation strategy to mitigate the annotation burden of high-quality instructions. EnvDrop~\cite{tan2019learning} employs environmental dropout to augment limited training data by mimicking unseen environments. 
Due to the success of Transformer-based cross-modal pretraining~\cite{li2020unicoder,li2020oscar,chen2020uniter,lu2019vilbert,Li2019VisualBERTAS}, recent works have explored transformer architectures into VLN tasks~\cite{hao2020towards,hong2021vln,Chen2021HistoryAM,Moudgil2021SOATAS,Qi2021TheRT,Guhur2021AirbertIP}. 
HAMT~\cite{Chen2021HistoryAM} develops a history aware multimodal transformer to better encode the long-horizon navigation history. DUET~\cite{Chen2022ThinkGA} constructs a dual-scale graph transformer for joint long-term action planning and fine-grained cross-modal understanding.
HOP~\cite{Qiao2022HOPHA} introduces a new history-and-order aware pretraining paradigm for encouraging the learning of spatio-temporal multimodal correspondence. 
However, these pretraining-based methods still learn to align the raw directional features and visual features trained using one-hot labels to the linguistic instruction features, leading to limited performance due to the large semantic gap among these multi-modal inputs. 

In contrast to the above mentioned VLN approaches, in this work, we  build a bridge among multi-modal inputs for facilitating the alignment by introducing actional atomic concepts formed by language. Through these  actional atomic concepts, the alignment can be significantly simplified and good interpretability can be provided.

\textbf{Contrastive Language-Image Pretraining (CLIP).}
CLIP~\cite{radford2021learning} is a cross-modal pretrained model using 400 million image and text pairs collected from the web. Through natural language supervision~\cite{Jia2021ScalingUV,Sariyildiz2020LearningVR,Desai2021VirTexLV} rather than one-hot labels of fixed size of object categories, CLIP has shown great potential in open-world object recognition. Recently, many works have attempted to introduce CLIP into various computer vision (CV) or vision-language (V\&L) tasks to improve the generalization ability of the downstream models~\cite{Song2022CLIPMA,Subramanian2022ReCLIPAS,Khandelwal2021SimpleBE,Shen2021HowMC,Rao2021DenseCLIPLD,Dai2022EnablingMG,Liang2022VisualLanguageNP}. DenseCLIP~\cite{Rao2021DenseCLIPLD} introduces CLIP into dense prediction tasks, e.g., semantic segmentation,  through converting the original image-text matching in CLIP to the pixel-text matching.~\cite{Dai2022EnablingMG} distills the vision-language knowledge learned in CLIP to enhance the multimodal generation models. EmbCLIP~\cite{Khandelwal2021SimpleBE} investigates the ability of CLIP's visual representations in improving embodied AI tasks. Some works have also tried to apply CLIP into VLN tasks~\cite{Liang2022VisualLanguageNP,Shen2021HowMC}. ProbES~\cite{Liang2022VisualLanguageNP} utilizes the knowledge learned from CLIP to build an in-domain dataset by self-exploration for pretraining.~\cite{Shen2021HowMC} replaces the ResNet visual encoder pretrained on ImageNet in the conventional VLN models with the pretrained CLIP visual encoder. 

In this paper, we resort to the powerful object recognition ability of CLIP to provide object concepts for each single-view observation. To encourage  instruction-oriented object concept extraction for better alignment, a concept refining adapter is further introduced beyond CLIP to re-rank its predicted object concepts according to the instruction. 

\begin{table*}[t]
	\fontsize{6}{6}\selectfont

\caption{Atomic Action Concept Mapping.}
	\vspace{-0.2cm}
	\label{tab:atomic action concept mapping}
	\resizebox{1.0\linewidth}{!}{
	{\renewcommand{\arraystretch}{1.2}
		\begin{tabular}{c||c|c|c|c|c|c|c}

			\specialrule{.1em}{.05em}{.05em}
			\multirow{2}{*}{Elevation $\tilde{\theta}_{t,n}$}&\multicolumn{7}{c}{Heading $\tilde{\psi}_{t,n}$ }\cr\cline{2-8}
			&(-2$\pi$, -3$\pi$/2]&(-3$\pi$/2, -$\pi$/2)&[-$\pi$/2, 0)&0&(0, $\pi$/2]&($\pi$/2, 3$\pi$/2)&[3$\pi$/2, 2$\pi$)\cr
			\hline
			$>$0&\multicolumn{7}{c}{go up}\cr\cline{2-8}
			$<$0&\multicolumn{7}{c}{go down}\cr\cline{2-8}
			0&turn right&go back&turn left&go forward&turn right&go back&turn left\\
        
 \specialrule{.1em}{.05em}{.05em}

			\end{tabular}}}
	\vspace{-0.4cm}
\end{table*}
\section{Preliminaries}
\subsection{VLN Problem Setup}
Given a natural language instruction $I=\{w_{1},...,w_{l}\}$ with $l$ words, a VLN agent is asked to navigate from a start viewpoint $S$ to the goal viewpoint $G$. At each timestep $t$, the agent receives a panoramic observation, containing $N_{o}$ image views  $O_{t}=\{O_{t,n}\}_{n=1}^{N_{o}}$. Each $O_{t,n}$ contains the image $B_{t,n}$ and the attached direction information $A_{t,n}$. The visual feature $\mathbf{v}_{t,n}$ for $B_{t,n}$ is obtained by a pretrained ResNet~\cite{he2016deep} or ViT~\cite{dosovitskiy2021an}. $A_{t,n}$ is usually composed of the heading $\psi_{t,n}$ and the elevation $\theta_{t,n}$. Each panoramic observation contains $d$ navigable viewpoints $C_{t}=\{C_{t,i}\}_{i=1}^{d}$ as the action candidates. At timestep $t$, the agent predicts an action $\mathbf{a}_{t}$ from $C_{t}$ based on the instruction $I$ and current visual observations $O_{t}$. 

\subsection{Baseline Agents}
Our AACL can be applied to many previous VLN models. In this work, two strong baseline agents HAMT~\cite{Chen2021HistoryAM} and DUET~\cite{Chen2022ThinkGA} are selected. In this section, we briefly describe one baseline HAMT. In HAMT, the agent receives the instruction $I$, the panoramic observation $O_{t}$, and the navigation history $H_{t}$ at each timestep $t$. $H_{t}$ is a sequence of historical visual observations. A standard BERT~\cite{Devlin2019BERTPO} is used to obtain the instruction feature $\mathbf{f}_{I}$ for $I$.
For each view $n$ with the angle information $<\psi_{t,n}, \theta_{t,n}>$ in $O_{t}$, the direction feature is defined by $\mathbf{e}_{A_{t,n}}=(\mathrm{sin}\psi_{t,n},\mathrm{cos}\psi_{t,n},\mathrm{sin}\theta_{t,n},\mathrm{cos}{\theta_{t,n}}$). With the visual feature $\mathbf{v}_{t,n}$ and the direction feature $\mathbf{e}_{A_{t,n}}$, the observation embedding $\mathbf{o}_{t,n}$ for each view $n$ is calculated by:
\begin{equation}
\label{eq:compose}
\begin{aligned}
\mathbf{o}_{t,n}&=\mathrm{Dr}(\mathrm{LN}(\mathrm{LN}(\mathbf{W}_{v}\mathbf{v}_{t,n})+\mathrm{LN}(\mathbf{W}_{a}\mathbf{e}_{A_{t,n}})\\
&+\mathbf{e}_{t,n}^{N}+\mathbf{e}_{v}^{T})),
\end{aligned}
\end{equation}
where $\mathrm{LN}(\cdot)$ and $\mathrm{Dr}(\cdot)$ denote layer normalization~\cite{Ba2016LayerN} and dropout, respectively. $\mathbf{W}_{v}$ and  $\mathbf{W}_{a}$  are learnable weights, and $\mathbf{e}_{t,n}^{N}$ and $\mathbf{e}_{v}^{T}$ denote the navigable embedding and the type embedding, respectively \cite{Chen2021HistoryAM}. The observation feature $\mathbf{o}_{t}$ is represented by $\mathbf{o}_{t}=[\mathbf{o}_{t,1};...;\mathbf{o}_{t,N_{o}}]$, where $N_{o}$ is the number of views. And a hierarchical vision transformer~\cite{Chen2021HistoryAM} is constructed to get the history feature $\mathbf{h}_{t}=[\mathbf{h}_{t,1};...;\mathbf{h}_{t,t-1}]$ for the navigation history $H_{t}$. 

Then $\mathbf{f}_{I}$, $\mathbf{o}_t$, and $\mathbf{h}_t$ are fed into a cross-modal transformer encoder $E^{c}(\cdot)$, resulting in:

\begin{equation}
\label{eq:ec}
\mathbf{\tilde{f}}_{I}^{t},\mathbf{\tilde{o}}_t,\mathbf{\tilde{h}}_t=E^{c}(\mathbf{f}_{I},[\mathbf{o}_t;\mathbf{h}_t]).
\end{equation}

The updated instruction feature $\mathbf{\tilde{f}}_{I}^{t}$ and observation feature $\mathbf{\tilde{o}}_t$ are used for action prediction:
\begin{equation}
\label{eq:ea}
\mathbf{a}_{t} = E^{a}(\mathbf{\tilde{f}}_{I}^{t}, \mathbf{\tilde{o}}_t),
\end{equation}
where $E^{a}(\cdot)$ is a two-layer fully-connected network. For more model details, refer to~\cite{Chen2021HistoryAM}.

In Eq.~\ref{eq:compose}, HAMT obtains the observation feature $\mathbf{o}_{t,n}$ directly by the pretrained visual feature $\mathbf{v}_{t,n}$ and the raw direction feature $\mathbf{e}_{A_{t,n}}$. 
In AACL, we map the observations $O_{t,n}$ to actional atomic concepts $U_{t,n}$ formed by language and use $U_{t,n}$ to obtain the new observation feature $\mathbf{o}'_{t,n}$. In this way,  the gap between $O_{t}$ and $I$ can be effectively mitigated to simplify the alignment.

\section{Actional Atomic-Concept Learning}
In this section, we describe our AACL in detail, the overview of which is presented in Figure~\ref{fig:overview}. At timestep $t$, the agent receives multi-modal inputs $I$,  $O_{t}$, and $H_{t}$ similar to HAMT. For each $O_{t,n}$ in $O_{t}$ containing the single-view image $B_{t,n}$ and the direction $A_{t,n}$, AACL first conducts object concept mapping and atomic action concept mapping to obtain the object concept $U_{t,n}^{obj}$ and the action concept $U_{t,n}^{act}$. And a concept refining adapter is built to re-rank $U_{t,n}^{obj}$ according to the instruction $I$ for better alignment. The actional atomic concept $U_{t,n}$ is then obtained by concatenating $U_{t,n}^{act}$ and $U_{t,n}^{obj}$, and fed to the concept encoder $E^{t}(\cdot)$ to get the concept feature $\mathbf{\tilde{u}}_{t,n}$. Finally, an observation co-embedding module is constructed to use $\mathbf{\tilde{u}}_{t,n}$ for regularizing the visual feature $\mathbf{v}_{t,n}$ and the directional feature $\mathbf{e}_{A_{t,n}}$ to get new observation features $\mathbf{o}'_{t,n}$. For $H_{t}$ which contains historical visual observations, we also use AACL to get the enhanced history features $\mathbf{h}'_{t}$ like $O_{t}$. Then $\mathbf{o}'_{t}=\{\mathbf{o}'_{t,n}\}_{n=1}^{N_{o}}$, $\mathbf{h}'_{t}$, and the instruction features $\mathbf{f}_{I}$ are fed to the cross-modal Transformer encoder $E^{c}(\cdot)$ for calculating the action predictions $\mathbf{a}'_{t}$. 

\subsection{Actional Atomic-Concept Mapping}
\label{Actional} 

\textbf{Object Concept Mapping.}
For each observation $O_{t,n}$ containing the single-view image $B_{t,n}$, we map $B_{t,n}$ to get the object concept $U_{t,n}^{obj}$  based on a pre-built in-domain object concept repository. Benefiting from large-scale language supervision from 400M image-text pairs, CLIP~\cite{radford2021learning} has more powerful open-world object recognition ability than conventional image classification or object detection models pretrained on a fixed-size object category set. In this work, we resort to CLIP to conduct the object concept mapping considering its good generalizability.  
Concretely, the object concept repository $\{U_{c}^{obj}\}_{c=1}^{N_{c}}$ is constructed by extracting object words from the training dataset, where $N_{c}$ is the repository size. And we get the image feature $\mathbf{f}_{B_{t,n}}$ through the pretrained CLIP Image Encoder $E^{v}_{\mathrm{CLIP}}(\cdot)$:
\begin{equation}
\mathbf{f}_{B_{t,n}}=E^{v}_{\mathrm{CLIP}}(B_{t,n}).
\end{equation}
For object concept $U_{c}^{obj}$, we construct the text phrase $T_{c}$ formed as ``a photo of a \{$U_{c}^{obj}$\}''. Then the text feature $\mathbf{f}_{T_{c}}$ is derived through the pretrained CLIP Text Encoder $E^{t}_{\mathrm{CLIP}}(\cdot)$:
\begin{equation}
\mathbf{f}_{T_{c}} =E^{t}_{\mathrm{CLIP}}(T_{c}).
\end{equation}
Then the mapping probability of the image $B_{t,n}$ regarding the object concept $U_{c}^{obj}$ is calculated by:
\begin{equation}
\mathbf{p}(y=U_{c}^{obj}|B_{t,n}) = \frac{\mathrm{exp}(\mathrm{sim}(\mathbf{f}_{B_{t,n}},\mathbf{f}_{T_{c}})/\tau)}{\sum_{c=1}^{N_{c}}(\mathrm{exp}(\mathrm{sim}(\mathbf{f}_{B_{t,n}},\mathbf{f}_{T_{c}})/\tau))},
\end{equation}
where $\mathrm{sim}(\cdot,\cdot)$  denotes the cosine similarity, $\tau$ represents the temperature parameter.  Considering that a single-view image in the observation usually contains more than one salient object, we extract the top $k$ object concepts (text) having the maximum mapping probabilities conditioned on  $B_{t,n}$ as its corresponding object concepts, i.e., $U_{t,n}^{obj}=\{U^{obj}_{t,n,i}\}_{i=1}^{k}$.

\textbf{Atomic Action Concept Mapping.}
The atomic action concept $U_{t,n}^{act}$ for $O_{t,n}$ can be derived through its  directional information $A_{t,n}$ and the directional information $\tilde{A}_{t-1}$ of the agent's selected action at timestep $t$-1. We first use six basic actions in VLN tasks to build the predefined atomic action set, i.e., {\it go up, go down, go forward, go back,  turn right,} and {\it turn left}. Denote $A_{t,n}=<\psi_{t,n},\theta_{t,n}>$, where $\psi_{t,n} \in [0,2\pi)$ and $\theta_{t,n} \in [-\frac{\pi}{2},\frac{\pi}{2}]$ are the heading and the elevation, respectively. Similarly, $\tilde{A}_{t-1}=<\tilde{\psi}_{t-1},\tilde{\theta}_{t-1}>$, where $\tilde{\psi}_{t-1} \in [0,2\pi)$ and $\tilde{\theta}_{t-1} \in [-\frac{\pi}{2},\frac{\pi}{2}]$. We calculate the relative direction of $<\psi_{t,n},\theta_{t,n}>$ to $<\tilde{\psi}_{t-1},\tilde{\theta}_{t-1}>$ by:
\begin{equation}
\begin{aligned}
\tilde{\psi}_{t,n}=\psi_{t,n}- \tilde{\psi}_{t-1}, \quad \tilde{\theta}_{t,n} =\theta_{t,n} -  \tilde{\theta}_{t-1}. 
\end{aligned}
\end{equation}
Then we use $<\tilde{\psi}_{t,n}, \tilde{\theta}_{t,n}>$ to obtain $U_{t,n}^{act}.$ Following the direction judgement rule in VLN~\cite{anderson2018vision}, we use $\tilde{\theta}_{t,n}$ first to judge whether $U^{act}_{t,n}$ is ``go up'' or ``go down'' by comparing it to zero. Otherwise, $U^{act}_{t,n}$ is further determined through $\tilde{\psi}_{t,n}$. Specifically, if $\tilde{\psi}_{t,n}$ is equal to zero, $U^{act}_{t,n}$ is    ``go forward''. Otherwise, $U^{act}_{t,n}$ is further determined to be ``turn right'', ``turn left'', or ``go back''. The detailed mapping rule is listed in Table~\ref{tab:atomic action concept mapping}.

\subsection{Concept Refining Adapter}
\label{Concept Refining Adapter}
After getting $\{U^{obj}_{t,n,i}\}_{i=1}^{k}$ and $U_{t,n}^{act}$ for each $O_{t,n}$, the actional atomic concept $\{U_{t,n,i}\}_{i=1}^{k}$ can be obtained by directly concatenating $U_{t,n}^{act}$ and each $U_{t,n,i}^{obj}$. 
A direct way to obtain the actional atomic concept feature $\mathbf{u}_{t,n}$  is to feed each $U_{t,n,i}$ to the concept encoder $E^{t}(\cdot)$ and get a weighted sum based on their object prediction probability $\mathbf{p}_{i}$ by CLIP:
\vspace{-0.2cm}
\begin{equation}
\label{eq:concept}
\mathbf{u}_{t,n}=\sum_{i=1}^{k}\mathbf{p}_{i} \cdot E^{t}(U_{t,n,i}), 
\end{equation}
where $E^{t}(\cdot)$ is the BERT-like language encoder. 
However, even if CLIP can extract informative object concepts for each observation, some noisy object concepts may exist and extracting instruction-oriented object concepts would be more useful for alignment and making action decisions.
Inspired by~\cite{Gao2021CLIPAdapterBV}, we propose to construct a concept refining adapter beyond CLIP to refine the object concept under the constraint of the instruction.  Given a feature $\mathbf{f}$, the concept refining adapter is written as:
\begin{equation}
A(\mathbf{f}) = \mathrm{ReLU}(\mathbf{f}^{T}\mathbf{W}_{1})\mathbf{W}_{2},
\end{equation}
where $\mathbf{W}_{1}$ and $\mathbf{W}_{2}$ are learnable parameters, and $\mathrm{ReLU}(\cdot)$ is the rectified linear unit for activation.  Denote the instruction feature as $\mathbf{f}_{I}=\{\mathbf{f}_{I}^{cls},\mathbf{f}_{I}^{1},...\mathbf{f}_{I}^{l}\}$. For the image feature $\mathbf{f}_{B_{t,n}}$ of the single-view image $B_{t,n}$, we obtain the updated image feature $\mathbf{\tilde{f}}_{B_{t,n}}$ by feeding  $\mathbf{f}_{B_{t,n}}$ and $\mathbf{f}_{I}^{cls}$ to $A(\cdot)$:
\begin{equation}
\label{adapter}
\mathbf{\tilde{f}}_{B_{t,n}} = \alpha \cdot \mathbf{f}_{B_{t,n}} + (1-\alpha) \cdot A([\mathbf{f}_{B_{t,n}};\mathbf{f}_{I}^{cls}]),
\end{equation}
where $\alpha$ servers as the residual ratio to help adjust the degree of maintaining the original knowledge for better performance~\cite{Gao2021CLIPAdapterBV}, and $[\cdot;\cdot]$ denotes feature concatenation. Denote the top $k$ object concept features obtained by CLIP for the single-view image $B_{t,n}$ as $\{\mathbf{f}_{T_{i}}\}_{i=1}^{k}$. We use the updated image feature $\mathbf{\tilde{f}}_{B_{t,n}}$ to get the re-ranking object prediction probability $\mathbf{\tilde{p}}$ of $\{\mathbf{f}_{T_{i}}\}_{i=1}^{k}$:
\begin{equation}
\mathbf{\tilde{p}} = \mathrm{Softmax}(\mathrm{sim}(\mathbf{\tilde{f}}_{B_{t,n}},\mathbf{f}_{T_{1}}),...,\mathrm{sim}(\mathbf{\tilde{f}}_{B_{t,n}},\mathbf{f}_{T_{k}})).\\
\end{equation}
Then we get the refined concept feature $\mathbf{\tilde{u}}_{t,n}$ by replacing $\mathbf{p}$ in Eq.~\ref{eq:concept} by $\mathbf{\tilde{p}}$.


\subsection{Observation Co-Embedding}
\label{Observation Co-embedding}
After obtaining the concept feature $\mathbf{\tilde{u}}_{t,n}$ for each single-view observation $O_{t,n}$, we introduce an observation co-embedding module to use $\mathbf{\tilde{u}}_{t,n}$ for bridging multi-modal inputs and calculating the final observation feature $\mathbf{o}'_{t,n}$.
At first, we separately embed the visual feature $\mathbf{v}_{t,n}$, the direction feature $\mathbf{e}_{A_{t,n}}$, and the concept feature $\mathbf{\tilde{u}}_{t,n}$ to obtain  $\mathbf{o}_{t,n}^{v}$,  $\mathbf{o}_{t,n}^{a}$, and $\mathbf{o}_{t,n}^{u}$, respectively, by:
\begin{equation}
\label{separate}
\begin{aligned}
\mathbf{o}_{t,n}^{v}&=\mathrm{Dr}(\mathrm{LN}(\mathrm{LN}(\mathbf{\tilde{W}}_{v}\mathbf{v}_{t,n})+\mathbf{e}_{t,n}^{N}+\mathbf{e}_{v}^{T})),\\
\mathbf{o}_{t,n}^{a}&=\mathrm{Dr}(\mathrm{LN}(\mathrm{LN}(\mathbf{\tilde{W}}_{a}\mathbf{e}_{A_{t,n}})+\mathbf{e}_{t,n}^{N}+\mathbf{e}_{v}^{T})),\\
\mathbf{o}_{t,n}^{u}&=\mathrm{Dr}(\mathrm{LN}(\mathrm{LN}(\mathbf{\tilde{W}}_{u}\mathbf{\tilde{u}}_{t,n})+\mathbf{e}_{t,n}^{N}+\mathbf{e}_{v}^{T})),
\end{aligned}
\end{equation}
where $\mathbf{\tilde{W}}_{v}$, $\mathbf{\tilde{W}}_{a}$ and $\mathbf{\tilde{W}}_{u}$ are learnable weights. 

Unlike HAMT that combining different features into one embedding (Eq.~\ref{eq:compose}), we keep the separate embeddings as in Eq.~\ref{separate} such that a new observation contrast strategy can be performed.
Concretely, the view embedding $\mathbf{o}_{t,n}^{v}$ and the direction embedding $\mathbf{o}_{t,n}^{a}$ are summed as the visual embedding $\mathbf{o}_{t,n}^{V}=\mathbf{o}_{t,n}^{v}+\mathbf{o}_{t,n}^{a}$. Then $\mathbf{o}_{t,n}^{V}$ in each single-view observation $O_{t,n}$ is forced to stay close to the paired concept embedding $\mathbf{o}_{t,n}^{u}$ while staying far away from the concept embeddings $\overline{\mathbf{o}}_{t,n}^{u}$ in other single-view observations in $O_{t}$: 
\begin{equation}
\mathcal{L}_{\mathrm{c}}=-\sum_{t}\sum_{n}\mathrm{log}(\frac{\mathrm{e}^{\mathrm{sim}(\mathbf{o}_{t,n}^{V}, \mathbf{o}_{t,n}^{u})/\tau}}{\mathrm{e}^{\mathrm{sim}(\mathbf{o}_{t,n}^{V}, \mathbf{o}_{t,n}^{u})/\tau}+\sum\limits_{\overline{\mathbf{o}}_{t,n}^{u}}\mathrm{e}^{\mathrm{sim}(\mathbf{o}_{t,n}^{V}, \overline{\mathbf{o}}_{t,n}^{u})/\tau}}),
\end{equation}
where $\tau$ is the temperature parameter. By observation contrast, the discrimination of each single-view observation can be effectively enhanced and the semantic gap between observations and instructions can be largely mitigated with the help of the actional atomic concept. 
To fully merge the information for each observation, we use $\mathbf{o}'_{t,n}=\mathbf{o}_{t,n}^{V}+\mathbf{o}_{t,n}^{u}$ to obtain the final observation feature $\mathbf{o}'_{t,n}$.  

\subsection{Action Prediction}
Similar to the observation feature $\mathbf{o}'_{t}=\{\mathbf{o}'_{t,n}\}_{n=1}^{N_{o}}$ ($N_{o}$ is the number of views), the history feature $\mathbf{h}'_{t}$ is obtained for $H_{t}$ through AACL. 
With $\mathbf{o}'_{t}$, $\mathbf{h}'_t$, and the instruction feature  $\mathbf{f}_{I}$, the action $\mathbf{a}'_{t}$ can be obtained from the cross-modal transformer encoder $E^{c}(\cdot)$ and the action prediction module $E^{a}(\cdot)$ (see Eq.~\ref{eq:ec} and Eq.~\ref{eq:ea}).
Following most existing VLN works~\cite{tan2019learning,hong2021vln,Chen2021HistoryAM}, we combine Imitation Learning (IL) and Reinforcement Learning (RL) to train VLN agents. Let the imitation learning loss be $\mathcal{L}_{\mathrm{IL}}$ and the reinforcement learning loss be $\mathcal{L}_{\mathrm{RL}}$. The total training objective of AACL is:
\begin{equation}
\mathcal{L}=\mathcal{L}_{\mathrm{RL}}+\lambda_{1}\mathcal{L}_{\mathrm{IL}}+\lambda_{2}\mathcal{L}_{\mathrm{c}},
\end{equation}
where $\lambda_{1}$ and $\lambda_{2}$ are balance parameters. 

\begin{table*}[!htb]
	\fontsize{7}{7}\selectfont

\caption{Comparison with the SOTA methods on R2R. }
	\vspace{-0.2cm}
	\label{tab:com with sota}
	\resizebox{1.0\linewidth}{!}{
	{\renewcommand{\arraystretch}{1.1}
		\begin{tabular}{c||c|c|c|c|c|c|c|c|c|c|c|c}

			\specialrule{.1em}{.05em}{.05em}
			\multirow{2}{*}{Method}&\multicolumn{4}{c|}{Val Seen }&\multicolumn{4}{c|}{Val Unseen}&\multicolumn{4}{c}{Test Unseen}\cr\cline{2-13}
			&TL&NE $\downarrow$&SR $\uparrow$&SPL $\uparrow$&TL&NE $\downarrow$&SR $\uparrow$&SPL $\uparrow$&TL&NE $\downarrow$&SR $\uparrow$&SPL $\uparrow$\cr
			\hline
			
        
            Seq2Seq~\cite{anderson2018vision}&11.33&6.01&39&-&8.39&7.81&22&-&8.13&7.85&20&18\\
            RCM+SIL(train) \cite{wang2019reinforced}&10.65&3.53&67&-&11.46&6.09&43&-&11.97&6.12&43&38\\
            
   
          EnvDropout~\cite{tan2019learning}&11.00&3.99&62&59&10.70&5.22&52&48&11.66&5.23&51&47\\  
                         
        PREVALENT~\cite{hao2020towards} & 10.32 & 3.67 & 69 & 65 & 10.19 & 4.71 & 58 & 53 & 10.51 & 5.30 & 54 & 51 \\
		 ORIST~\cite{Qi2021TheRT}&-&-&-&-&10.90&4.72&57&51&11.31&5.10&57&52\\ VLN$\circlearrowright$BERT~\cite{hong2021vln}&11.13&2.90&72&68&12.01&3.93&63&57&12.35&4.09&63&57\\
 HOP~\cite{Qiao2022HOPHA}&11.51&2.46&76&70&12.52&3.79&64&57&13.29&3.87&64&58\\
		 \hline HAMT~\cite{Chen2021HistoryAM} (baseline)&11.15&2.51&76&72&11.46&3.62&66&61&12.27&3.93&65&60\\ 
     HAMT+AACL (ours)&11.31&2.53&76&72&12.09&3.41&69&\textbf{63}&12.74&3.71&66&\textbf{61}\\
    DUET~\cite{Chen2022ThinkGA} (baseline)&12.32&2.28&79&\textbf{73}&13.94&3.31&72&60&14.73&3.65&69&59\\
    DUET+AACL (ours)&13.32&\textbf{2.15}&\textbf{80}&72&15.01&\textbf{3.00}&\textbf{74}&61&15.47&\textbf{3.38}&\textbf{71}&59\\
 \specialrule{.1em}{.05em}{.05em}

		\end{tabular}}}
	\vspace{-0.2cm}
\end{table*}

\begin{table*}[!htb]
	\fontsize{20}{20}\selectfont

\caption{Navigation and object grounding performance on REVERIE.}
	\vspace{-0.2cm}
	\label{tab:com with sota on reverie}
	\resizebox{1.0\linewidth}{!}{
	{\renewcommand{\arraystretch}{1.2}
		\begin{tabular}{c||c|c|c|c|c|c|c|c|c|c|c|c|c|c|c|c|c|c}

			\specialrule{.1em}{.05em}{.05em}
			\multirow{2}{*}{Method}&\multicolumn{6}{c|}{Val Seen}&\multicolumn{6}{c|}{Val Unseen}&\multicolumn{6}{c}{Test Unseen}\cr\cline{2-19}
			&TL&SR $\uparrow$&OSR$\uparrow$&SPL $\uparrow$&RGS$\uparrow$&RGSPL$\uparrow$&TL&SR $\uparrow$&OSR$\uparrow$&SPL $\uparrow$&RGS$\uparrow$&RGSPL$\uparrow$&TL&SR $\uparrow$&OSR$\uparrow$&SPL $\uparrow$&RGS$\uparrow$&RGSPL$\uparrow$\cr
			\hline
			
        
            RCM~\cite{wang2019reinforced}&10.70&23.33&29.44&21.82&16.23&15.36&11.98&9.29&14.23&6.97&4.89&3.89&10.60&7.84&11.68&6.67&3.67&3.14\\
           
          SMNA~\cite{ma2019self}&7.54&41.25&43.29&39.61&30.07&28.98&9.07&8.15&11.28&6.44&4.54&3.61&9.23&5.80&8.39&4.53&3.10&2.39\\
         FAST-MATTN~\cite{qi2020reverie}&16.35&50.53&55.17&45.50&31.97&29.66&45.28&14.40&28.20&7.19&7.84&4.67&39.05&19.88&30.63&11.60&11.28&6.08\\ 
    SIA~\cite{Lin2021SceneIntuitiveAF}&13.61&61.91&65.85&57.08&45.96&42.65&41.53&31.53&44.67&16.28&22.41&11.56&48.61&30.80&44.56&14.85&19.02&9.20\\
            VLN$\circlearrowright$BERT~\cite{hong2021vln}&13.44&51.79&53.90&47.96&38.23&35.61&16.78&30.67&35.02&24.90&18.77&15.27&15.86&29.61&32.91&23.99&16.50&13.51\\
    HOP~\cite{Qiao2022HOPHA}&14.05&54.81&56.08&48.05&40.55&35.79&17.16&30.39&35.30&25.10&18.23&15.31&17.05&29.12&32.26&23.37&17.13&13.90\\
 \hline HAMT~\cite{Chen2021HistoryAM} (baseline)&12.79&43.29&47.65&40.19&27.20&25.18&14.08&32.95&36.84&30.20&18.92&17.28&13.62&30.40&33.41&26.67&14.88&13.08\\

    HAMT+AACL (ours)&13.01&42.52&46.66&39.48&28.39&26.48&14.08&34.17&38.54&29.70&20.53&17.69&13.30&35.52&39.57&31.34&18.04&15.96\\
    DUET~\cite{Chen2022ThinkGA} (baseline)&13.86&71.75&73.86&63.94&57.41&51.14&22.11&46.98&51.07&\textbf{33.73}&32.15&\textbf{23.03}&21.30&52.51&56.91&36.06&31.88&22.06\\
    DUET+AACL (ours)&14.54&\textbf{74.63}&\textbf{76.67}&\textbf{66.04}&\textbf{59.38}&\textbf{52.57}&\textbf{23.77}&\textbf{49.42}&\textbf{53.93}&33.54&\textbf{33.31}&22.49&21.88&\textbf{55.09}&\textbf{59.92}&\textbf{37.08}&\textbf{33.17}&\textbf{22.55}\\
 \specialrule{.1em}{.05em}{.05em}

			\end{tabular}}}
	\vspace{-0.4cm}
\end{table*}

\section{Experiments}
\subsection{Experimental Setup}
\label{Experimental Setup}
\textbf{Datasets.} We evaluate the proposed AACL on several popular VLN benchmarks with both fine-grained instructions (R2R~\cite{anderson2018vision}) and high-level instructions (REVERIE~\cite{qi2020reverie} and R2R-Last~\cite{Chen2021HistoryAM}). R2R includes 90 photo-realistic indoor scenes with 7189 trajectories. The dataset is split into train, val seen, val unseen, and test unseen sets with 61, 56, 11, and 18 scenes, respectively. REVERIE replaces the fine-grained instructions in R2R with high-level instructions which mainly target at object localization. R2R-Last only uses the last sentence of the original R2R instruction as the instruction.

\begin{table}

    \caption{Comparison on R2R-Last.}
    \vspace{-0.2cm}
    \small
    \centering
    \resizebox{\linewidth}{!}{
    \begin{tabular}{l|cccc}
    \specialrule{.1em}{.05em}{.05em}
			\multirow{2}{*}{Method}&\multicolumn{2}{c|}{Val Seen}&\multicolumn{2}{c}{Val Unseen}\cr\cline{2-5}
			&SR$\uparrow$&SPL$\uparrow$&SR$\uparrow$&SPL$\uparrow$\cr
			\hline
			
        EnvDrop~\cite{tan2019learning}&42.8&38.4&34.3&28.3\\
        VLN$\circlearrowright$BERT~\cite{hong2021vln}&50.2&45.8&41.6&37.3\\
        \hline
        HAMT~\cite{Chen2021HistoryAM} (baseline)&53.3&50.3&45.2&41.2\\
        HAMT+AACL (ours)&\textbf{54.2}&\textbf{51.1}&\textbf{47.2}&\textbf{42.1}\\
 \specialrule{.1em}{.05em}{.05em}
    \end{tabular}
}
\label{tab:com on r2rlast}
\vspace{-0.2cm}
\end{table}

\begin{table}
\fontsize{4}{4}\selectfont
    \caption{Ablation Study on R2R. The baseline agent we choose is HAMT~\cite{Chen2021HistoryAM}.}
\vspace{-0.2cm}
    \small
    \centering
    \resizebox{0.7\linewidth}{!}{
    {\renewcommand{\arraystretch}{1.0}
    \begin{tabular}{l|ccc}
    \specialrule{.1em}{.05em}{.05em}
			\multirow{2}{*}{Method}&\multicolumn{3}{c}{Val Unseen}\cr\cline{2-4}
   &NE$\downarrow$&SR$\uparrow$&SPL$\uparrow$\cr

			\hline
			
        separate embedding&3.66&66.67&61.19\\
       w/o contrast&3.42&67.94&61.48\\ 
       w/o refine&3.45&67.82&62.33\\
            full model&\textbf{3.41}&\textbf{68.54}&\textbf{62.96}\\
        
 \specialrule{.1em}{.05em}{.05em}
    \end{tabular}
}}
\label{tab:ablation}
\vspace{-0.4cm}
\end{table}

\textbf{Evaluation Metrics.} We adopt the common metrics used in previous works~\cite{Chen2021HistoryAM,anderson2018vision,qi2020reverie} to evaluate the model performance: 1) Navigation Error (NE) calculates the average distance between the agent stop position and the target viewpoint, 2) Trajectory Length (TL) is the average path length in meters, 3) Success Rate (SR) is the ratio of stopping within 3 meters to the goal, 4) Success rate weighted by Path Length (SPL) makes the trade-off between SR and TL, 5) Oracle Success Rate (OSR) calculates the ratio of containing a viewpoint along the path where the target object is visible, 6) Remote Grounding Success Rate (RGS) is the ratio of performing correct object grounding when stopping, and 7) Remote Grounding Success weighted by Path Length (RGSPL) weights RGS by TL. 1)--4), 3)--4), and 2)--7) are used for evaluation on R2R, R2R-Last, and REVERIE, respectively. 

\textbf{Baselines.} 
In this work, we choose two strong baseline agents, HAMT~\cite{Chen2021HistoryAM} and DUET~\cite{Chen2022ThinkGA} to verify AACL's effectiveness. In HAMT, a hierarchical transformer is adopted for storing historical observations and actions. In contrast, DUET keeps track of all visited and navigable locations through a topological map. 

\textbf{Implementation Details.} 
We implement our model using the MindSpore Lite tool~\cite{mindspore}.
The batch size is set to 8, 8, 4 on R2R, R2R-Last, and REVERIE, respectively. The temperature parameter $\tau$ is set to 0.5. 
The loss weight $\lambda_{1}$ is set to 0.2 on all datasets, and the loss weight $\lambda_{2}$ is set to 1, 1, and 0.01 on R2R, REVERIE, and R2R-Last, respectively.
The residual ratio in Eq.~\ref{adapter} is set to 0.8 empirically. During object concept mapping, we remain top 5 object predictions for each observation. 
The learning rate of the concept refining adapter is set to 0.1.  

\begin{figure*}[!htb]
\begin{centering}
\includegraphics[width=0.98\linewidth]{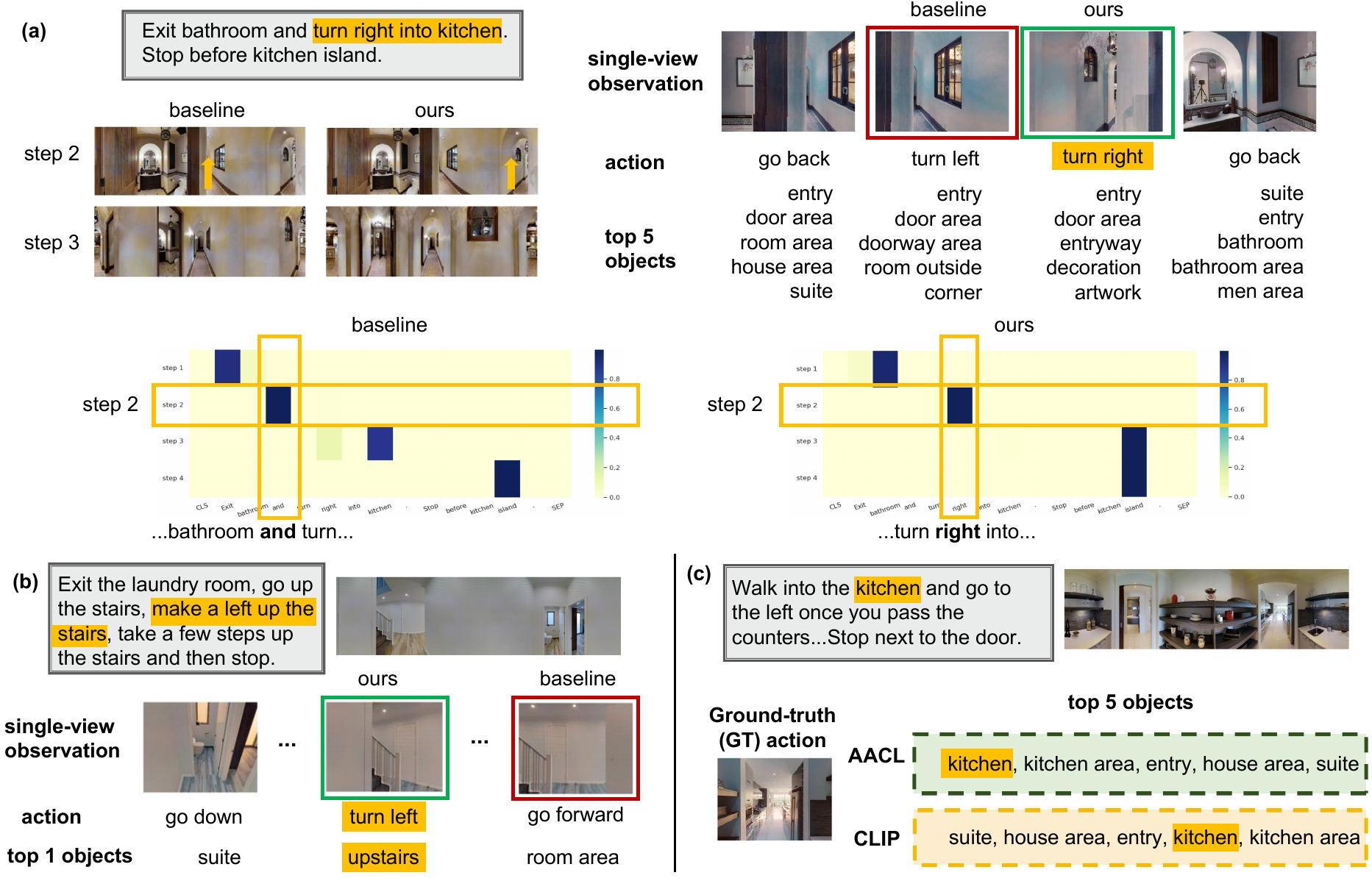}
\par\end{centering}
\caption{Visualization examples of action selection ((a) and (b)) and object concept mapping ((c)). In (a) and (b), the baseline is HAMT~\cite{Chen2021HistoryAM}. The green boxes denote the correct actions and the red boxes denote the wrong ones.
}
\vspace{-0.6cm}
\label{fig:visualization_3}
\end{figure*}

\subsection{Quantitative Results}

\textbf{Comparison with the State-of-the-Arts (SOTAs).} Table~\ref{tab:com with sota}\footnote{The original value 2.29 of NE under Val Unseen in HAMT is a typo, which is actually 3.62 confirmed by the author.}, Table~\ref{tab:com with sota on reverie}, and Table~\ref{tab:com on r2rlast} present the performance comparison between the SOTA methods and AACL, where we can find that AACL establishes new state-of-the-art results in most metrics on R2R, REVERIE and R2R-Last. These results show that AACL is useful not only when the instructions are fine-grained but also when the instruction information is limited, demonstrating that the proposed actional atomic concepts can effectively enhance the observation features, simplify their alignment to the linguistic instruction features, and therefore improve the agent performance. Moreover, we can find that AACL consistently outperforms the two strong baselines on these three benchmarks especially under Unseen scenarios, showing that AACL can be used as a general tool for the multi-modal alignment. 

\textbf{Ablation Study.} Table~\ref{tab:ablation} gives the ablation study of AACL. ``separate embedding'' means using the separate embedding scheme (Eq.~\ref{separate}) only for the visual feature and the directional feature. By comparing the results between ``separate embedding'' and ``w/o contrast'', we can find that the direct introduction of actional atomic concepts under the separate embedding strategy can already improve the navigation performance (1.27\% increase on SR), showing their effectiveness for enhancing the observation features. By comparing the results between ``w/o contrast'' and ``w/o refine'', we can observe that the proposed observation contrast strategy can effectively regularize the observation representation and improve the performance (0.85\% increase on SPL). The comparison between ``w/o refine'' and ``full model'' further shows the effectiveness of the concept refining adapter, demonstrating that the instruction-oriented object concept extraction can facilitate better cross-modal alignment.

\subsection{Qualitative Results}
Figure~\ref{fig:visualization_3} visualizes some results of action decision and object concept mapping. We can find that by introducing the actional atomic concepts, the agent is able to perform better cross-modal alignment for improving action decisions. In Figure~\ref{fig:visualization_3}(a), although the candidate observations do not contain the visual appearance of ``kitchen'', with the help of the actional atomic concepts,  AACL successfully chooses the right action whose paired action concept matches the one mentioned in the instruction (``turn right''), while the baseline selects the wrong one. From the  instruction attention comparison in the lower part of Figure~\ref{fig:visualization_3}(a), we can also observe that AACL attends to the right part (``right'') of the instruction at step 2, while the baseline attends to the wrong part (``and'').  In Figure~\ref{fig:visualization_3}(b), within the actional atomic concept, AACL successfully chooses the correct action asked in the instruction. In Figure~\ref{fig:visualization_3}(c), we can observe that the probability of ``kitchen'' of AACL is higher than that of CLIP for the GT action (top-1 vs. top-4), showing that the concept refining adapter enables more instruction-oriented object concept extraction, which is useful for selecting correct actions. 
\section{Conclusion}
In this work, we propose Actional Atomic-Concept Learning, which helps VLN agents demystify the alignment in VLN tasks through actional atomic concepts formed by language. 
During navigation, each visual observation is mapped to the specific actional atomic concept through the VLN environment and CLIP. A concept refining adapter is constructed to enable the instruction-oriented concept extraction. An observation co-embedding module is introduced to use concept features to regularize observation features.  
Experiments on public
VLN benchmarks show that AACL achieves new SOTA results. Benefiting from these human-understandable actional atomic concepts, AACL shows excellent interpretability in making action decision. 



\clearpage

\section{Acknowledgements}
This work was supported in part by National Key R\&D Program of China under Grant No. 2020AAA0109700, National Natural Science Foundation of China (NSFC) under Grant No.61976233, Guangdong Outstanding Youth Fund (Grant No. 2021B1515020061), Guangdong Natural Science Foundation under Grant 2017A030312006, Shenzhen Fundamental Research Program (Project No. RCYX20200714114642083, No. JCYJ20190807154211365), and the Fundamental Research Funds for the Central Universities, Sun Yat-sen University under Grant No. 22lgqb38, CAAI-Huawei MindSpore Open Fund. 
We gratefully acknowledge the support of MindSpore, CANN (Compute Architecture for Neural Networks), and Ascend AI Processor used for this research.

\bibliography{aaai23.bib} 

\begin{thebibliography}{45}
\providecommand{\natexlab}[1]{#1}

\bibitem[{{Anderson} et~al.(2018){Anderson}, {Wu}, {Teney}, {Bruce}, {Johnson}, {Sunderhauf}, {Reid}, {Gould}, and van~den {Hengel}}]{anderson2018vision}
{Anderson}, P.; {Wu}, Q.; {Teney}, D.; {Bruce}, J.; {Johnson}, M.; {Sunderhauf}, N.; {Reid}, I.; {Gould}, S.; and van~den {Hengel}, A. 2018.
\newblock Vision-and-Language Navigation: Interpreting Visually-Grounded Navigation Instructions in Real Environments.
\newblock In \emph{CVPR}.

\bibitem[{Ba, Kiros, and Hinton(2016)}]{Ba2016LayerN}
Ba, J.; Kiros, J.~R.; and Hinton, G.~E. 2016.
\newblock Layer Normalization.
\newblock \emph{ArXiv}, abs/1607.06450.

\bibitem[{Chen et~al.(2019)Chen, Suhr, Misra, Snavely, and Artzi}]{Chen2019TOUCHDOWNNL}
Chen, H.; Suhr, A.; Misra, D.~K.; Snavely, N.; and Artzi, Y. 2019.
\newblock TOUCHDOWN: Natural Language Navigation and Spatial Reasoning in Visual Street Environments.
\newblock In \emph{CVPR}.

\bibitem[{Chen et~al.(2021)Chen, Guhur, Schmid, and Laptev}]{Chen2021HistoryAM}
Chen, S.; Guhur, P.-L.; Schmid, C.; and Laptev, I. 2021.
\newblock History Aware Multimodal Transformer for Vision-and-Language Navigation.
\newblock In \emph{NeurIPS}.

\bibitem[{Chen et~al.(2022)Chen, Guhur, Tapaswi, Schmid, and Laptev}]{Chen2022ThinkGA}
Chen, S.; Guhur, P.-L.; Tapaswi, M.; Schmid, C.; and Laptev, I. 2022.
\newblock Think Global, Act Local: Dual-scale Graph Transformer for Vision-and-Language Navigation.
\newblock In \emph{CVPR}.

\bibitem[{{Chen} et~al.(2020){Chen}, {Li}, {Yu}, {Kholy}, {Ahmed}, {Gan}, {Cheng}, and {Liu}}]{chen2020uniter}
{Chen}, Y.-C.; {Li}, L.; {Yu}, L.; {Kholy}, A.~E.; {Ahmed}, F.; {Gan}, Z.; {Cheng}, Y.; and {Liu}, J. 2020.
\newblock UNITER: UNiversal Image-TExt Representation Learning.
\newblock In \emph{ECCV}.

\bibitem[{Dai et~al.(2022)Dai, Hou, Shang, Jiang, Liu, and Fung}]{Dai2022EnablingMG}
Dai, W.; Hou, L.; Shang, L.; Jiang, X.; Liu, Q.; and Fung, P. 2022.
\newblock Enabling Multimodal Generation on CLIP via Vision-Language Knowledge Distillation.
\newblock In \emph{ACL}.

\bibitem[{Desai and Johnson(2021)}]{Desai2021VirTexLV}
Desai, K.; and Johnson, J. 2021.
\newblock VirTex: Learning Visual Representations from Textual Annotations.
\newblock In \emph{CVPR}.

\bibitem[{Devlin et~al.(2019)Devlin, Chang, Lee, and Toutanova}]{Devlin2019BERTPO}
Devlin, J.; Chang, M.-W.; Lee, K.; and Toutanova, K. 2019.
\newblock BERT: Pre-training of Deep Bidirectional Transformers for Language Understanding.
\newblock \emph{ArXiv}, abs/1810.04805.

\bibitem[{{Dosovitskiy} et~al.(2021){Dosovitskiy}, {Beyer}, {Kolesnikov}, {Weissenborn}, {Zhai}, {Unterthiner}, {Dehghani}, {Minderer}, {Heigold}, {Gelly}, {Uszkoreit}, and {Houlsby}}]{dosovitskiy2021an}
{Dosovitskiy}, A.; {Beyer}, L.; {Kolesnikov}, A.; {Weissenborn}, D.; {Zhai}, X.; {Unterthiner}, T.; {Dehghani}, M.; {Minderer}, M.; {Heigold}, G.; {Gelly}, S.; {Uszkoreit}, J.; and {Houlsby}, N. 2021.
\newblock An Image is Worth 16x16 Words: Transformers for Image Recognition at Scale.
\newblock In \emph{ICLR}.

\bibitem[{{Fried} et~al.(2018){Fried}, {Hu}, {Cirik}, {Rohrbach}, {Andreas}, {Morency}, {Berg-Kirkpatrick}, {Saenko}, {Klein}, and {Darrell}}]{fried2018speaker}
{Fried}, D.; {Hu}, R.; {Cirik}, V.; {Rohrbach}, A.; {Andreas}, J.; {Morency}, L.-P.; {Berg-Kirkpatrick}, T.; {Saenko}, K.; {Klein}, D.; and {Darrell}, T. 2018.
\newblock Speaker-Follower Models for Vision-and-Language Navigation.
\newblock In \emph{NeurIPS}.

\bibitem[{{Fu} et~al.(2020){Fu}, {Wang}, {Peterson}, {Grafton}, {Eckstein}, and {Wang}}]{fu2020counterfactual}
{Fu}, T.-J.; {Wang}, X.~E.; {Peterson}, M.~F.; {Grafton}, S.~T.; {Eckstein}, M.~P.; and {Wang}, W.~Y. 2020.
\newblock Counterfactual Vision-and-Language Navigation via Adversarial Path Sampler.
\newblock In \emph{ECCV}.

\bibitem[{Gao et~al.(2021)Gao, Geng, Zhang, Ma, Fang, Zhang, Li, and Qiao}]{Gao2021CLIPAdapterBV}
Gao, P.; Geng, S.; Zhang, R.; Ma, T.; Fang, R.; Zhang, Y.; Li, H.; and Qiao, Y.~J. 2021.
\newblock CLIP-Adapter: Better Vision-Language Models with Feature Adapters.
\newblock \emph{ArXiv}, abs/2110.04544.

\bibitem[{Guhur et~al.(2021)Guhur, Tapaswi, Chen, Laptev, and Schmid}]{Guhur2021AirbertIP}
Guhur, P.-L.; Tapaswi, M.; Chen, S.; Laptev, I.; and Schmid, C. 2021.
\newblock Airbert: In-domain Pretraining for Vision-and-Language Navigation.
\newblock In \emph{ICCV}.

\bibitem[{{Hao} et~al.(2020){Hao}, {Li}, {Li}, {Carin}, and {Gao}}]{hao2020towards}
{Hao}, W.; {Li}, C.; {Li}, X.; {Carin}, L.; and {Gao}, J. 2020.
\newblock Towards Learning a Generic Agent for Vision-and-Language Navigation via Pre-Training.
\newblock In \emph{CVPR}.

\bibitem[{{He} et~al.(2016){He}, {Zhang}, {Ren}, and {Sun}}]{he2016deep}
{He}, K.; {Zhang}, X.; {Ren}, S.; and {Sun}, J. 2016.
\newblock Deep Residual Learning for Image Recognition.
\newblock In \emph{CVPR}.

\bibitem[{{Hong} et~al.(2021){Hong}, {Wu}, {Qi}, {Rodriguez-Opazo}, and {Gould}}]{hong2021vln}
{Hong}, Y.; {Wu}, Q.; {Qi}, Y.; {Rodriguez-Opazo}, C.; and {Gould}, S. 2021.
\newblock VLN BERT: A Recurrent Vision-and-Language BERT for Navigation.
\newblock In \emph{CVPR}.

\bibitem[{Jia et~al.(2021)Jia, Yang, Xia, Chen, Parekh, Pham, Le, Sung, Li, and Duerig}]{Jia2021ScalingUV}
Jia, C.; Yang, Y.; Xia, Y.; Chen, Y.-T.; Parekh, Z.; Pham, H.; Le, Q.~V.; Sung, Y.-H.; Li, Z.; and Duerig, T. 2021.
\newblock Scaling Up Visual and Vision-Language Representation Learning With Noisy Text Supervision.
\newblock In \emph{ICML}.

\bibitem[{Khandelwal et~al.(2022)Khandelwal, Weihs, Mottaghi, and Kembhavi}]{Khandelwal2021SimpleBE}
Khandelwal, A.; Weihs, L.; Mottaghi, R.; and Kembhavi, A. 2022.
\newblock Simple but Effective: CLIP Embeddings for Embodied AI.
\newblock In \emph{CVPR}.

\bibitem[{{Ku} et~al.(2020){Ku}, {Anderson}, {Patel}, {Ie}, and {Baldridge}}]{ku2020room}
{Ku}, A.; {Anderson}, P.; {Patel}, R.; {Ie}, E.; and {Baldridge}, J. 2020.
\newblock Room-Across-Room: Multilingual Vision-and-Language Navigation with Dense Spatiotemporal Grounding.
\newblock In \emph{EMNLP}.

\bibitem[{{Li} et~al.(2020{\natexlab{a}}){Li}, {Duan}, {Fang}, {Gong}, and {Jiang}}]{li2020unicoder}
{Li}, G.; {Duan}, N.; {Fang}, Y.; {Gong}, M.; and {Jiang}, D. 2020{\natexlab{a}}.
\newblock Unicoder-VL: A Universal Encoder for Vision and Language by Cross-Modal Pre-Training.
\newblock In \emph{AAAI}.

\bibitem[{Li et~al.(2019)Li, Yatskar, Yin, Hsieh, and Chang}]{Li2019VisualBERTAS}
Li, L.~H.; Yatskar, M.; Yin, D.; Hsieh, C.-J.; and Chang, K.-W. 2019.
\newblock VisualBERT: A Simple and Performant Baseline for Vision and Language.
\newblock \emph{ArXiv}, abs/1908.03557.

\bibitem[{{Li} et~al.(2020{\natexlab{b}}){Li}, {Yin}, {Li}, {Zhang}, {Hu}, {Zhang}, {Wang}, {Hu}, {Dong}, {Wei}, {Choi}, and {Gao}}]{li2020oscar}
{Li}, X.; {Yin}, X.; {Li}, C.; {Zhang}, P.; {Hu}, X.; {Zhang}, L.; {Wang}, L.; {Hu}, H.; {Dong}, L.; {Wei}, F.; {Choi}, Y.; and {Gao}, J. 2020{\natexlab{b}}.
\newblock Oscar: Object-Semantics Aligned Pre-training for Vision-Language Tasks.
\newblock In \emph{ECCV}.

\bibitem[{Liang et~al.(2022)Liang, Zhu, Li, Xu, and Liang}]{Liang2022VisualLanguageNP}
Liang, X.; Zhu, F.; Li, L.; Xu, H.; and Liang, X. 2022.
\newblock Visual-Language Navigation Pretraining via Prompt-based Environmental Self-exploration.
\newblock In \emph{ACL}.

\bibitem[{Lin, Li, and Yu(2021)}]{Lin2021SceneIntuitiveAF}
Lin, X.; Li, G.; and Yu, Y. 2021.
\newblock Scene-Intuitive Agent for Remote Embodied Visual Grounding.
\newblock In \emph{CVPR}.

\bibitem[{{Lu} et~al.(2019){Lu}, {Batra}, {Parikh}, and {Lee}}]{lu2019vilbert}
{Lu}, J.; {Batra}, D.; {Parikh}, D.; and {Lee}, S. 2019.
\newblock ViLBERT: Pretraining Task-Agnostic Visiolinguistic Representations for Vision-and-Language Tasks.
\newblock In \emph{NeurIPS}.

\bibitem[{{Ma} et~al.(2019){Ma}, jiasen {lu}, {Wu}, {AlRegib}, {Kira}, richard {socher}, and {Xiong}}]{ma2019self}
{Ma}, C.-Y.; jiasen {lu}; {Wu}, Z.; {AlRegib}, G.; {Kira}, Z.; richard {socher}; and {Xiong}, C. 2019.
\newblock Self-Monitoring Navigation Agent via Auxiliary Progress Estimation.
\newblock In \emph{ICLR}.

\bibitem[{MindSpore(2022)}]{mindspore}
MindSpore. 2022.
\newblock MindSpore.
\newblock \url{https://www.mindspore.cn/}.

\bibitem[{Mnih et~al.(2016)Mnih, Badia, Mirza, Graves, Lillicrap, Harley, Silver, and Kavukcuoglu}]{Mnih2016AsynchronousMF}
Mnih, V.; Badia, A.~P.; Mirza, M.; Graves, A.; Lillicrap, T.~P.; Harley, T.; Silver, D.; and Kavukcuoglu, K. 2016.
\newblock Asynchronous Methods for Deep Reinforcement Learning.
\newblock In \emph{ICML}.

\bibitem[{Moudgil et~al.(2021)Moudgil, Majumdar, Agrawal, Lee, and Batra}]{Moudgil2021SOATAS}
Moudgil, A.; Majumdar, A.; Agrawal, H.; Lee, S.; and Batra, D. 2021.
\newblock SOAT: A Scene- and Object-Aware Transformer for Vision-and-Language Navigation.
\newblock In \emph{NeurIPS}.

\bibitem[{Nguyen and Daum{\'e}(2019)}]{Nguyen2019HelpAV}
Nguyen, K.; and Daum{\'e}, H. 2019.
\newblock Help, Anna! Visual Navigation with Natural Multimodal Assistance via Retrospective Curiosity-Encouraging Imitation Learning.
\newblock In \emph{EMNLP}.

\bibitem[{Qi et~al.(2021)Qi, Pan, Hong, Yang, van~den Hengel, and Wu}]{Qi2021TheRT}
Qi, Y.; Pan, Z.; Hong, Y.; Yang, M.-H.; van~den Hengel, A.; and Wu, Q. 2021.
\newblock The Road to Know-Where: An Object-and-Room Informed Sequential BERT for Indoor Vision-Language Navigation.
\newblock In \emph{ICCV}.

\bibitem[{Qi et~al.(2020)Qi, Pan, Zhang, van~den Hengel, and Wu}]{Qi2020ObjectandActionAM}
Qi, Y.; Pan, Z.; Zhang, S.; van~den Hengel, A.; and Wu, Q. 2020.
\newblock Object-and-Action Aware Model for Visual Language Navigation.
\newblock In \emph{ECCV}.

\bibitem[{{Qi} et~al.(2020){Qi}, {Wu}, {Anderson}, {Wang}, {Wang}, {Shen}, and van~den {Hengel}}]{qi2020reverie}
{Qi}, Y.; {Wu}, Q.; {Anderson}, P.; {Wang}, X.; {Wang}, W.~Y.; {Shen}, C.; and van~den {Hengel}, A. 2020.
\newblock REVERIE: Remote Embodied Visual Referring Expression in Real Indoor Environments.
\newblock In \emph{CVPR}.

\bibitem[{Qiao et~al.(2022)Qiao, Qi, Hong, Yu, Wang, and Wu}]{Qiao2022HOPHA}
Qiao, Y.; Qi, Y.; Hong, Y.; Yu, Z.; Wang, P.; and Wu, Q. 2022.
\newblock HOP: History-and-Order Aware Pre-training for Vision-and-Language Navigation.
\newblock In \emph{CVPR}.

\bibitem[{{Radford} et~al.(2021){Radford}, {Kim}, {Hallacy}, {Ramesh}, {Goh}, {Agarwal}, {Sastry}, {Askell}, {Mishkin}, {Clark}, {Krueger}, and {Sutskever}}]{radford2021learning}
{Radford}, A.; {Kim}, J.~W.; {Hallacy}, C.; {Ramesh}, A.; {Goh}, G.; {Agarwal}, S.; {Sastry}, G.; {Askell}, A.; {Mishkin}, P.; {Clark}, J.; {Krueger}, G.; and {Sutskever}, I. 2021.
\newblock Learning Transferable Visual Models From Natural Language Supervision.
\newblock In \emph{ICML}.

\bibitem[{Rao et~al.(2022)Rao, Zhao, Chen, Tang, Zhu, Huang, Zhou, and Lu}]{Rao2021DenseCLIPLD}
Rao, Y.; Zhao, W.; Chen, G.; Tang, Y.; Zhu, Z.; Huang, G.; Zhou, J.; and Lu, J. 2022.
\newblock DenseCLIP: Language-Guided Dense Prediction with Context-Aware Prompting.
\newblock In \emph{CVPR}.

\bibitem[{Sariyildiz, Perez, and Larlus(2020)}]{Sariyildiz2020LearningVR}
Sariyildiz, M.~B.; Perez, J.; and Larlus, D. 2020.
\newblock Learning Visual Representations with Caption Annotations.
\newblock In \emph{ECCV}.

\bibitem[{Shen et~al.(2022)Shen, Li, Tan, Bansal, Rohrbach, Chang, Yao, and Keutzer}]{Shen2021HowMC}
Shen, S.; Li, L.~H.; Tan, H.; Bansal, M.; Rohrbach, A.; Chang, K.-W.; Yao, Z.; and Keutzer, K. 2022.
\newblock How Much Can CLIP Benefit Vision-and-Language Tasks?
\newblock In \emph{ICLR}.

\bibitem[{Song et~al.(2022)Song, Dong, Zhang, Liu, and Wei}]{Song2022CLIPMA}
Song, H.; Dong, L.; Zhang, W.; Liu, T.; and Wei, F. 2022.
\newblock CLIP Models are Few-Shot Learners: Empirical Studies on VQA and Visual Entailment.
\newblock In \emph{ACL}.

\bibitem[{Subramanian et~al.(2022)Subramanian, Merrill, Darrell, Gardner, Singh, and Rohrbach}]{Subramanian2022ReCLIPAS}
Subramanian, S.; Merrill, W.; Darrell, T.; Gardner, M.; Singh, S.; and Rohrbach, A. 2022.
\newblock ReCLIP: A Strong Zero-Shot Baseline for Referring Expression Comprehension.
\newblock In \emph{ACL}.

\bibitem[{{Tan}, {Yu}, and {Bansal}(2019)}]{tan2019learning}
{Tan}, H.; {Yu}, L.; and {Bansal}, M. 2019.
\newblock Learning to Navigate Unseen Environments: Back Translation with Environmental Dropout.
\newblock In \emph{NAACL-HLT}.

\bibitem[{Wang, Wu, and Shen(2020)}]{Wang2020SoftER}
Wang, H.; Wu, Q.; and Shen, C. 2020.
\newblock Soft Expert Reward Learning for Vision-and-Language Navigation.
\newblock In \emph{ECCV}.

\bibitem[{{Wang} et~al.(2019){Wang}, {Huang}, {Celikyilmaz}, {Gao}, {Shen}, {Wang}, {Wang}, and {Zhang}}]{wang2019reinforced}
{Wang}, X.; {Huang}, Q.; {Celikyilmaz}, A.; {Gao}, J.; {Shen}, D.; {Wang}, Y.-F.; {Wang}, W.~Y.; and {Zhang}, L. 2019.
\newblock Reinforced Cross-Modal Matching and Self-Supervised Imitation Learning for Vision-Language Navigation.
\newblock In \emph{CVPR}.

\bibitem[{{Zhu} et~al.(2020){Zhu}, {Zhu}, {Chang}, and {Liang}}]{zhu2020vision}
{Zhu}, F.; {Zhu}, Y.; {Chang}, X.; and {Liang}, X. 2020.
\newblock Vision-Language Navigation With Self-Supervised Auxiliary Reasoning Tasks.
\newblock In \emph{CVPR}.

\end{thebibliography}

\appendix

\section{Appendix}

\subsection{Training Objectives}
In this section, we describe the imitation learning (IL) loss and the reinforcement learning loss (RL) used in training. Denote the predicted action at timestep $t$ as $\mathbf{a}_{t}$. The IL loss $\mathcal{L}_{\mathrm{IL}}$ can be calculated by:
\begin{equation}
\mathcal{L}_{\mathrm{IL}}=\sum_{t}-\mathbf{a}_{t}^{*}\mathrm{log}(\mathbf{a}_{t}),
\end{equation}
where $\mathbf{a}_{t}^{*}$ is the teacher action of the ground-truth path at timestep $t$. And the RL loss is formulated as:
\begin{equation}
\mathcal{L}_{\mathrm{RL}}=\sum_{t}-\mathbf{a}_{t}^{s}\mathrm{log}(\mathbf{a}_{t})A_{t},
\end{equation}
where $\mathbf{a}_{t}^{s}$ is the sampled action from the agent action prediction $\mathbf{a}_{t}$, and $A_{t}$ is the advantage calculated by A2C algorithm~\cite{Mnih2016AsynchronousMF}.

\subsection{Implementation Details}
\label{Implementation Details}
Following~\cite{Chen2021HistoryAM}, we also add an additional all-zero feature as the ``stop'' feature to the observation features during each timestep, and its paired  atomic action concept is set to {\it stop} manually. Since the observation contrast strategy is introduced to enhance the discrimination of the action candidate for pursuing accurate  action decision, it is not conducted for the navigation history feature. We construct the object concept repository by extracting object words from the training dataset and the augmentation dataset~\cite{hong2021vln}  of R2R for ensuring the diversity of the object concepts. During the object concept mapping, we directly use the CLIP visual feature (ViT-B/32) released by~\cite{Chen2021HistoryAM}. And the CLIP text encoder is fixed. The dimensionalities of the learnable parameters $\mathbf{W}_{1}$ and $\mathbf{W}_{2}$ in the concept refining adapter are set to 256 and 512, respectively. 

\subsection{More Quantitative Results}
In this section, we give more quantitative results in Table~\ref{tab:comparison on object number} and Table~\ref{tab:comparison w random}.  Table~\ref{tab:comparison on object number} present the comparison of different object number $k$ in object concept mapping. where we can find that ``$k$=5'' can achieve best performance in most metrics. This is reasonable because the number of salient objects in a single-view observation is usually not too much. Table~\ref{tab:comparison w random} investigates the results when using different kinds of  concepts for regularizing observation features. From Table~\ref{tab:comparison w random}, we can observe that using action and object concepts as actional atomic concepts is more for useful for improving the alignment and the navigation performance than using only object or action concepts.

\subsection{More Visualization Results}
\label{More Visualization Results}
\begin{figure*}[t]
\begin{centering}
\includegraphics[width=0.98\linewidth]{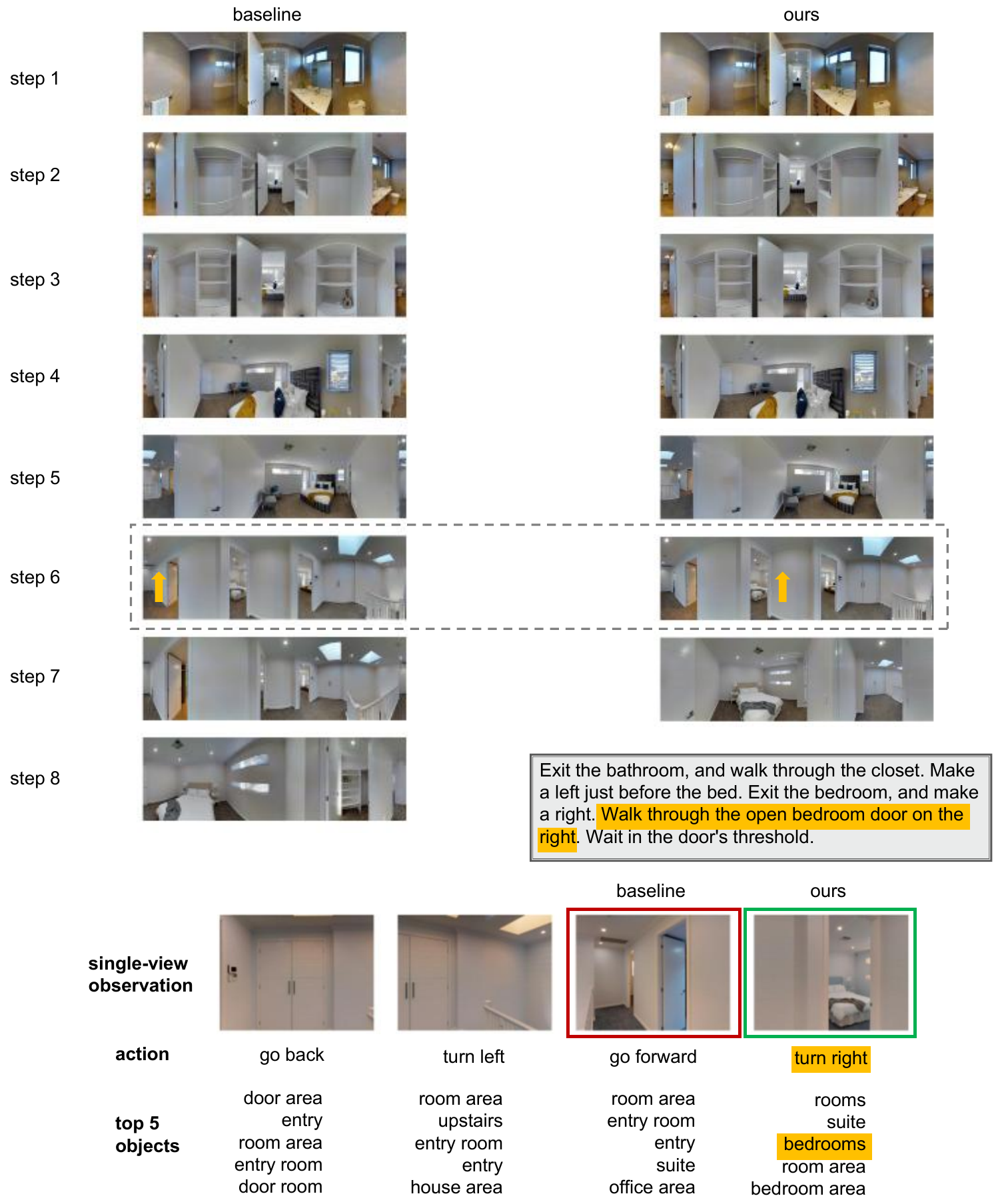}
\par\end{centering}
\caption{Visualization of the action selections by the baseline method~\cite{Chen2021HistoryAM} and our AACL. The green boxes denote the correct actions and the red boxes denote the wrong ones. After step 6 marked with the grey dashed box, the baseline and AACL make different trajectories.
}
\vspace{-0.4cm}
\label{fig:visualization 1}
\end{figure*}

\begin{figure*}[t]
\begin{centering}
\includegraphics[width=0.98\linewidth]{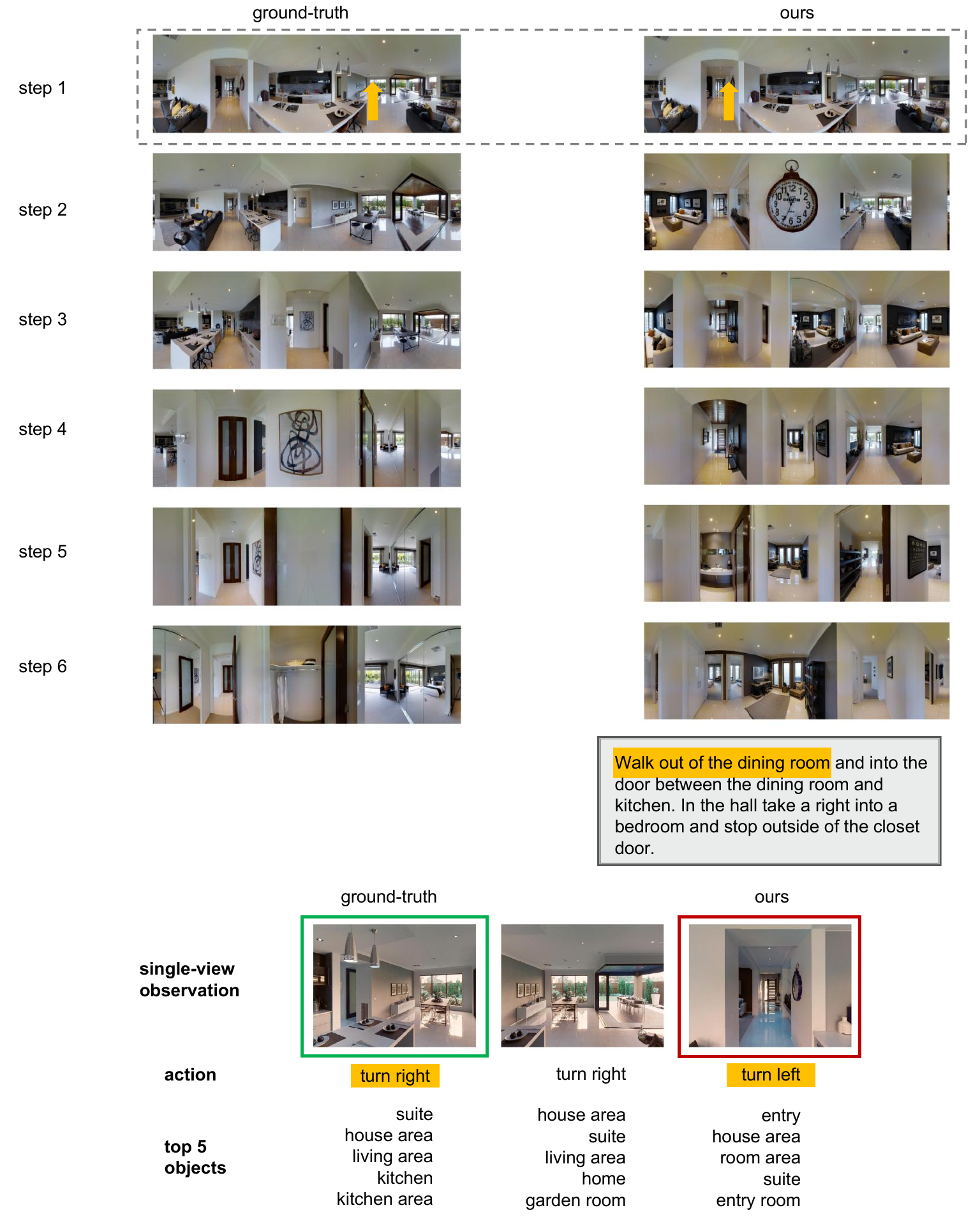}
\par\end{centering}
\caption{Failure case of AACL. The green boxes denote the correct actions and the red boxes denote the wrong ones. After step 1 marked with the grey dashed box, the ground-truth and AACL have different trajectories.
}
\vspace{-0.4cm}
\label{fig:failurecase 2}
\end{figure*}

\begin{figure*}[t]
\begin{centering}
\includegraphics[width=0.98\linewidth]{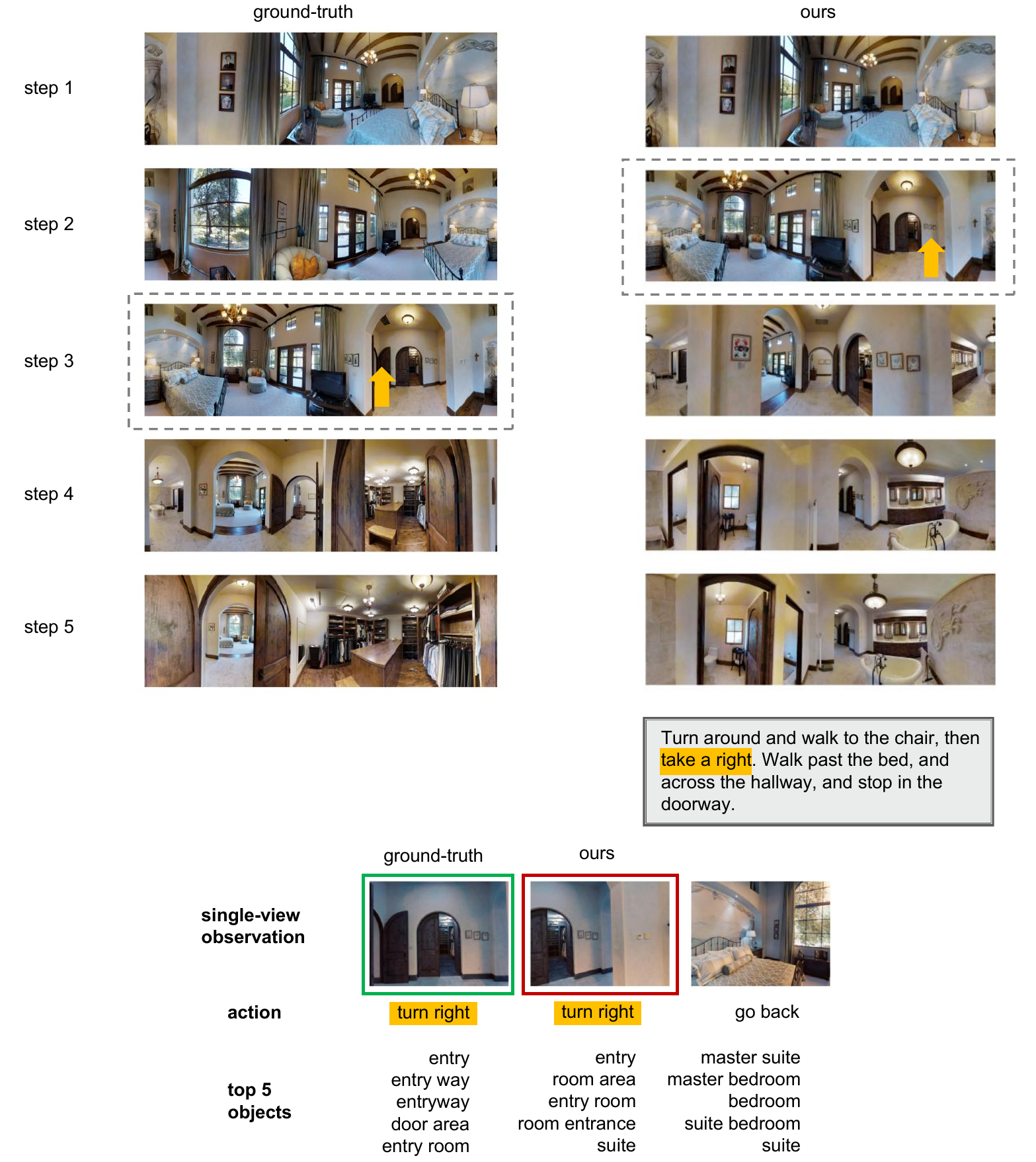}
\par\end{centering}
\caption{Failure case of AACL. The green boxes denote the correct actions and the red boxes denote the wrong ones. After the panoramic views marked with the grey dashed boxes, the ground-truth and AACL have different trajectories.
}
\vspace{-0.4cm}
\label{fig:failurecase 1}
\end{figure*}

In this section, we first give more visualization results of the action decisions by the baseline method~\cite{Chen2021HistoryAM} and AACL in Figure~\ref{fig:visualization 1} $\sim$ Figure~\ref{fig:failurecase 1}. From Figure~\ref{fig:visualization 1}, we can find that through AACL, the agent can better align the observation to the instruction to make correct action decision.  
In Figure~\ref{fig:visualization 1}, the correct action chosen by AACL contains the actional atomic concept of ``turn right bedrooms'', which matches the key information mentioned in the sub-instruction ``walk through the open bedroom door on the right''.   Figure~\ref{fig:failurecase 2} and Figure~\ref{fig:failurecase 1} present two failure cases. From  Figure~\ref{fig:failurecase 2}, we can find that although the action chosen by AACL is wrong, it can also be seen as a right action of ``walk out of the dining room'' due to the instruction ambiguity.  In Figure~\ref{fig:failurecase 1}, although the action chosen by AACL is different from the ground-truth one, its paired actional atomic concept is similar with that of the ground-truth action.  

\begin{figure*}[t]
\begin{centering}
\includegraphics[width=0.98\linewidth]{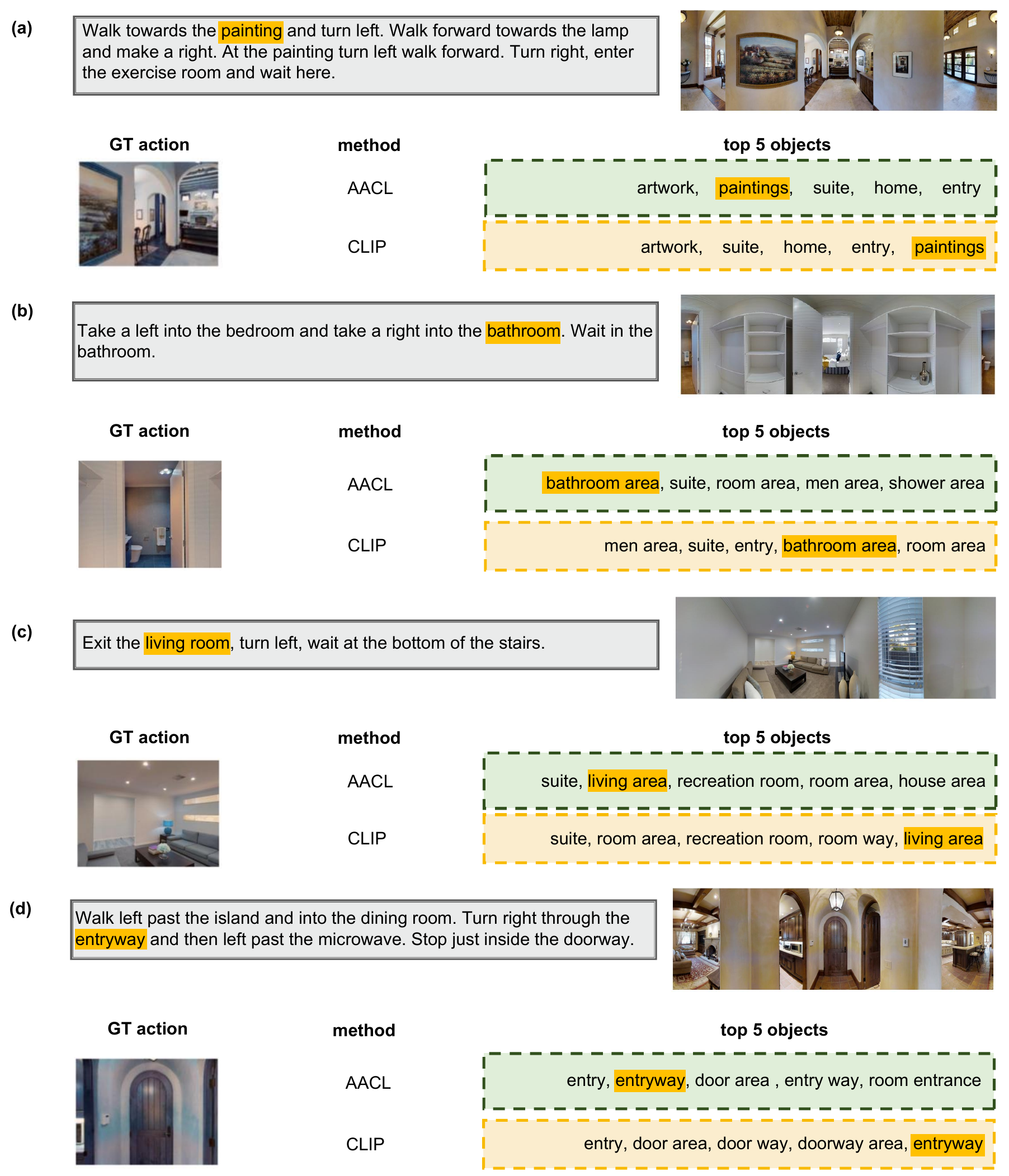}
\par\end{centering}
\caption{Visualization of the object predictions by CLIP~\cite{radford2021learning} and our AACL.
}
\vspace{-0.4cm}
\label{fig:visualization_2}
\end{figure*}

Figure~\ref{fig:visualization_2} gives more visualization results of the object predictions of CLIP~\cite{radford2021learning} and our AACL. We can find that the concept refining adapter can contribute to more  instruction-oriented object concept extraction. For example, in Figure~\ref{fig:visualization_2}(b), the probability of ``bathroom area'' of AACL is higher than that of CLIP for the ground-truth action candidate (top-1 vs. top-4). These visualization results show the effectiveness of the proposed concept refining adapter.

\begin{table}[t]
\centering
\caption{Comparison of different object numbers.}
\vspace{-0.2cm}
\resizebox{0.6\linewidth}{!}{{
    {\renewcommand{\arraystretch}{1.0}
    \begin{tabular}{l|ccc}
        \specialrule{.1em}{.05em}{.05em}
			Setting&NE$\downarrow$&SR$\uparrow$&SPL$\uparrow$\cr
    	\hline
			
        $k$=1&3.52&67.52&\textbf{62.24}\\
        $k$=5&\textbf{3.42}&\textbf{67.94}&61.48\\
        $k$=10&3.47&66.88&60.99\\
        
 \specialrule{.1em}{.05em}{.05em}
    \end{tabular}}}
}
\label{tab:comparison on object number}
\end{table}
\begin{table}[t]
\centering
\caption{Comparison of different kinds of  concepts.}
\vspace{-0.2cm}
\resizebox{0.7\linewidth}{!}{{
    {\renewcommand{\arraystretch}{1.0}
    \begin{tabular}{c|ccc}
        \specialrule{.1em}{.05em}{.05em}
			Setting&NE$\downarrow$&SR$\uparrow$&SPL$\uparrow$\cr

    	\hline
			 object&3.60&66.37&61.39\\
   action&3.51&67.13&61.34\\
   action+object&\textbf{3.42}&\textbf{67.94}&\textbf{61.48}\\
        
 \specialrule{.1em}{.05em}{.05em}
    \end{tabular}}}
}
\label{tab:comparison w random}
\vspace{-0.4cm}
\end{table}

\end{document}